\documentclass[10pt,twocolumn,letterpaper]{article}

\usepackage{sty}
\usepackage{times}
\usepackage{epsfig}
\usepackage{graphicx}
\usepackage{amsmath}
\usepackage{amssymb}
\usepackage{multirow}
\usepackage{booktabs} 
\usepackage{tabularx}
\usepackage{caption}
\usepackage{subcaption}

\usepackage{pifont}
\newcommand{\cmark}{\ding{51}}%
\newcommand{\xmark}{\ding{55}}

 \usepackage{xcolor}
\usepackage{url}

\usepackage{csquotes}
\usepackage{float}
\usepackage{bm}

\usepackage{amsfonts}       
\usepackage{nicefrac}       
\usepackage{microtype}      

\usepackage{soul}
\usepackage{rotating}

\newcolumntype{?}[0]{!{\vrule width 0.05mm}}

\usepackage{dsfont}

\newcommand\blfootnote[1]{%
  \begingroup
  \renewcommand\thefootnote{}\footnote{#1}%
  \addtocounter{footnote}{-1}%
  \endgroup
}

\usepackage[breaklinks=true,bookmarks=false]{hyperref}

\vcvcfinalcopy 

\begin{document}

\title{DILF-EN framework for Class-Incremental Learning}

\author{Mohammed Asad Karim$^{1,\ast}$,\hspace{0.5cm} Indu Joshi$^{2,\ast,a}$,\hspace{0.5cm} Pratik Mazumder$^{3,\ast}$, \hspace{0.5cm} Pravendra Singh$^{4,\ast}$\\
$^1$Independent Researcher, India\hspace{0.5cm}
$^2$Inria Sophia Antipolis, France\hspace{0.5cm}
$^3$IIT Kanpur, India\\
$^4$IIT Roorkee, India\\
{\tt\small asadkarim0938@gmail.com, indu.joshi@inria.fr, pratikm@cse.iitk.ac.in,}\\ {\tt\small pravendra.singh@cs.iitr.ac.in}
}

\maketitle

\ifvcvcfinal\thispagestyle{empty}\fi

\begin{abstract} 
A class-incremental learning problem is characterized by training data becoming available in a phase-by-phase manner. Deep learning models suffer from catastrophic forgetting of the classes in the older phases as they get trained on the classes introduced in the new phase. In this work, we show that the effect of catastrophic forgetting on the model prediction varies with the change in orientation of the same image, which is a novel finding. Based on this, we propose a novel data-ensemble approach that combines the predictions for the different orientations of the image to help the model retain further information regarding the previously seen classes and thereby reduce the effect of forgetting on the model predictions. However, we cannot directly use the data-ensemble approach if the model is trained using traditional techniques. Therefore, we also propose a novel dual-incremental learning framework that involves jointly training the network with two incremental learning objectives, i.e., the class-incremental learning objective and our proposed data-incremental learning objective. In the dual-incremental learning framework, each image belongs to two classes, i.e., the image class (for class-incremental learning) and the orientation class (for data-incremental learning). In class-incremental learning, each new phase introduces a new set of classes, and the model cannot access the complete training data from the older phases. In our proposed data-incremental learning, the orientation classes remain the same across all the phases, and the data introduced by the new phase in class-incremental learning acts as new training data for these orientation classes. We empirically demonstrate that the dual-incremental learning framework is vital to the data-ensemble approach. We apply our proposed approach to state-of-the-art class-incremental learning methods and empirically show that our framework significantly improves the performance of these methods. Our proposed method significantly improves the performance of the state-of-the-art method (AANets) on the CIFAR-100 dataset by absolute margins of 3.30\%, 4.28\%, 3.55\%, 4.03\%, for the number of phases P=50, 25, 10, and 5, respectively, which establishes the efficacy of the proposed work.
\end{abstract}

\section{Introduction} 
\blfootnote{$^\ast$ Equal contribution. All authors have contributed equally to this work.}
\blfootnote{$^a$ Work done by the author at IIT Delhi, India.}
Deep models are trained with an implicit assumption that the entire training data is available before the training is initiated. However, sometimes the complete training data might not be available initially, and new classes may become sequentially available for training the model. This problem setting is referred to as \textit{incremental} learning \cite{rebuffi2017icarl,itaml2020}. In this setting, training data becomes available in phases, where each new phase contains training images from a set of classes not seen earlier. However, as the model is trained on images of a new phase, the knowledge acquired from previously seen classes is overridden. This causes the \textit{catastrophic forgetting} of classes seen in the older phases \cite{rebuffi2017icarl,hou2019learning,wu2019large}. To avoid catastrophic forgetting~\cite{mccloskey1989catastrophic}, it is desired that a model should be able to incrementally update itself to learn new classes over time while preserving the knowledge of the older classes seen during the earlier phases.

\par Several approaches have been presented in the literature to reduce catastrophic forgetting of deep models. A common approach to reduce catastrophic forgetting is through the replay of data seen in the older phases \cite{rebuffi2017icarl,hou2019learning,castro2018end,douillard2020podnet}. Generative models have also been exploited to synthesize training data which can be used for the replay \cite{kamra2017deep,shin2017continual,venkatesan2017strategy}. Another approach to reduce catastrophic forgetting is through the regularization of the model parameters such that the model does not forget the knowledge acquired during the older phases \cite{rebuffi2017icarl,castro2018end,aljundi2018memory,kirkpatrick2017overcoming,li2017learning,lopez2017gradient,douillard2020podnet}. Other techniques for incremental learning include modifying the model architecture for new phases \cite{li2019learn,yoon2018lifelong} or exploiting a sub-network for each phase \cite{fernando2017evolution,golkar2019continual}.

In this work, we empirically demonstrate that the effect of catastrophic forgetting on the model predictions varies with different orientations of the same image. Different from the approaches proposed so far, this paper proposes to reduce the impact of catastrophic forgetting in state-of-the-art class-incremental learning methods using a novel \textit{data-ensemble} (EN) approach that utilizes this novel finding. Specifically, the data-ensemble approach applies transformations to any given test image in order to create versions of the same test image with different orientations. Then, the trained model is used to obtain the predictions produced for each orientation of the given test image. Since the effect of forgetting will vary with the orientation of the image, the data-ensemble approach combines the predictions for each orientation of the given test image to help the model retain additional information regarding the previously seen classes. This reduces the effect of catastrophic forgetting on the model predictions.

However, the data-ensemble approach cannot be directly applied to a model trained using the traditional techniques. In fact, we demonstrate in Sec.~\ref{sec:ablnodilf} that applying data-ensemble to such a model severely degrades its performance. Therefore, we also propose a novel dual-incremental learning framework (DILF) that jointly trains the network using two incremental learning objectives: the \textit{class-incremental} learning objective and our proposed \textit{data-incremental} learning objective. The proposed dual-incremental learning framework considers each image to belong to two classes, i.e., the image class and the orientation class. The model is trained to identify the image classes through the class-incremental objective, while it is trained to identify the orientation classes through the data-incremental objective. We perform various ablation experiments in Sec.~\ref{orientation_classes} to validate the orientation classes used in our method. We empirically demonstrate that the data-ensemble approach significantly improves the class-incremental learning performance of the model after it is trained using the dual-incremental learning framework (see Sec.~\ref{sec:ablnodilf}). We provide a detailed discussion on the motivations for our DILF-EN approach in Secs.~\ref{sec:motivation_en},\ref{sec:motivation_dilf}.

\par \textbf{Research Contributions:} Our work is the first to show that the effect of catastrophic forgetting on the model predictions varies on changing the orientation of the same image. This paper also presents a novel class-incremental learning approach DILF-EN that consists of a novel \textit{dual-incremental learning} framework (DILF) for training the network and a novel \textit{data-ensemble} (EN) approach for evaluating the network in the class-incremental learning setting. To the best of our knowledge, this is the first work to use the data-ensemble approach and a dual-incremental learning objective to improve the class-incremental learning performance of the network. We introduce DILF-EN in four recent state-of-the-art class-incremental learning methods \cite{hou2019learning,douillard2020podnet,liu2021adaptive,liu2020mnemonics}, and all of them achieve superior performance after the introduction of the proposed framework. Extensive experiments on standard datasets and different ablation studies are conducted in order to gain insights into the proposed DILF-EN approach.

\section{Related Work}
Incremental learning \cite{aljundi2017expert,chen2018lifelong,li2019online,delange2021continual,lopez2017gradient,gdumb2020,cauwenberghs2001incremental,kuzborskij2013n,mensink2013distance,ruping2001incremental} is broadly categorized as \textit{task-incremental} and \textit{class-incremental}. In task-incremental learning, for each phase, the model is trained on a new task where each task has a separate group of classes. During inference, the model is provided with a task-id, and the model performance is evaluated only on classes pertaining to that task \cite{chaudhry2018riemannian,chaudhry2018efficient,davidson2020sequential,shin2017continual,riemer2018learning,singh2021rectification,singh2020calibrating}. However, in class-incremental learning, task-id is not available during inference, and the model has to learn a unified classifier that classifies all the classes seen across the different phases of training \cite{rebuffi2017icarl,hou2019learning,wu2019large,castro2018end,hu2021distilling,liu2020mnemonics}. Therefore, class-incremental learning is a far more challenging problem than task-incremental learning. We now present the literature on incremental learning, which can be categorized as:

\textbf{Regularization based approaches:} These approaches ensure that while learning the classes introduced in a new phase, the model does not change drastically and remembers the classes introduced in the older phases. Li and Hoiem \cite{li2017learning} propose knowledge distillation to constrain the model. Hou \etal \cite{hou2019learning} propose a less-forget constraint along with a combination of regularizers to balance the magnitude of weights while increasing the inter-class separation. Tao \etal \cite{tao2020topology} propose a topology-preserving loss for maintaining topology in feature space. Yu \etal \cite{yu2020semantic} estimate semantic drift of features of older classes while training the model for learning new classes. Recently, Douillard \etal \cite{douillard2020podnet} propose a spatial-distillation loss to learn representations that help the model to prevent catastrophic forgetting. The authors in \cite{POMPONI2020139} perform regularization of the internal embeddings of the network in order to prevent catastrophic forgetting.

\textbf{Replay based approaches:} The methods that follow the replay based approach for incremental learning store some training data or data representations from earlier phases. This data is then used for replay or prototype rehearsal which helps the model to prevent catastrophic forgetting of older classes while learning the new ones. Rebuffi \etal \cite{rebuffi2017icarl} propose that the nearest neighbors of the average samples per class should be used to prepare the subset for replay, and they also use the distillation loss. Belouadah and Popescu \cite{belouadah2019il2m} propose sharing both exemplars and class statistics from older phases. Recently, authors in \cite{liu2020mnemonics} propose to parameterize exemplars and optimize them in an end-to-end fashion. Recently, Liu \etal \cite{liu2021adaptive} introduce stability and plasticity preserving residual blocks to prevent catastrophic forgetting. Some approaches use generative models \cite{goodfellow2014generative} to generate samples for replay \cite{kamra2017deep,shin2017continual,venkatesan2017strategy}.

\textbf{Parameter isolation based approaches:} These approaches assume no constraints on the model size and reserve different model parameters for different phases. These methods add new branches for new tasks while freezing older branches to avoid overwriting of model parameters. Rusu \etal \cite{rusu2016progressive} propose to define a new network for a new phase and conduct knowledge transfer from older models. Abati \etal \cite{abati2020conditional} propose to introduce task-specific gating modules. Rajasegaran \etal \cite{rajasegaran2019random} propose a dynamic path selection algorithm that enables the model to select optimal path for each new phase. Xu and Zhu \cite{xu2018reinforced} propose reinforcement learning for selecting the optimal architecture to learn each new task. The work in \cite{SOKAR20211} trains sparse deep neural networks and compresses the sparse connections of each task in the network. The work in \cite{WEI20191} expands the network for new tasks while preserving previous task knowledge and incorporating the topology and attribute of network nodes.

The above-mentioned approaches have been mainly applied to the incremental image classification problem. Researchers have also proposed incremental learning methods to address other types of real-world problems. The authors in \cite{LIU2020125} propose an incremental learning framework to perform facial landmark tracking. The work in \cite{SINGH202283} proposes a cancelable template generation framework to address the incremental biometric template generation problem.

Different from these approaches, we propose a DILF-EN framework to address the class-incremental learning problem that first trains the network using a novel dual-class incremental learning objective and evaluates the model using a novel data-ensemble approach. Our approach augments other class-incremental learning approaches and significantly improves their performance.

\section{Proposed Method}
\label{proposed}

\begin{table*}[t]
    \centering
                \caption{Table reports the number of correct predictions for 5000 test images from the base classes transformed to the orientation classes $o_1$ or $o_2$. The Table also reports the number of correct predictions obtained using the data-ensemble (EN) approach that combines the predictions for two versions of each test image with orientation classes $o_1$ and $o_2$, respectively. For this experiment, we evaluate the AANets model after it is trained on all the phases of CIFAR-100 $P=5$ using DILF. EN denotes our proposed data-ensemble approach. \cmark $o_1$ and \cmark $o_2$ denotes the number of correctly classified test images having orientation $o_1$ and $o_2$, respectively. \xmark $o_1$ and \xmark $o_2$ denotes the number of incorrectly classified test images having orientation $o_1$ and $o_2$, respectively. \cmark EN denotes the correctly classified test images obtained using the data-ensemble method. \textbf{\cmark $o_1$\cmark $o_2$} denotes the number of test examples that are correctly classified by the model for both $o_1$ and $o_2$ orientations. \textbf{\cmark EN\cmark $o_1$\cmark $o_2$} denotes the number of test examples that are correctly classified by the model for both $o_1$ and $o_2$ orientations and also classified correctly using our proposed data-ensemble (EN) approach. \textbf{\cmark $o_1$\xmark $o_2$} denotes the number of test examples that are correctly classified by the model only when they have the orientation $o_1$ and not when they have the orientation $o_2$. \textbf{\xmark $o_1$\cmark $o_2$} denotes the number of test examples that are correctly classified by the model only when they have the orientation $o_2$ and not when they have the orientation $o_1$.}
     \label{tab:difforget}
     \scalebox{1}{
\begin{tabular}{c|c|c|c|c|c|c}
    \toprule
        \textbf{\cmark $o_1$}  & \textbf{\cmark $o_2$} & \textbf{\cmark $o_1$\cmark $o_2$} & \textbf{\cmark $o_1$\xmark $o_2$} & \textbf{\xmark $o_1$\cmark $o_2$} & \textbf{\cmark EN} & \textbf{\cmark EN\cmark $o_1$\cmark $o_2$}\\
        \midrule
        3365 & 3312 & 2893 & 472 & 419 & 3478 & 2893\\

        \bottomrule
    \end{tabular}
    }
\end{table*}

\subsection{Motivation for Data-Ensemble}\label{sec:motivation_en}
The class-incremental learning problem is formalized as training a deep model on a sequence of phases, where in each phase, the model is provided with new training images from a set of new classes. The objective is to obtain a model that performs well on all the classes seen till the current training phase $p$.

We perform an experiment using our proposed DILF to study the effect of catastrophic forgetting on different orientations of the same image. Table~\ref{tab:difforget} reports the correct predictions by a model for two orientations ($o_1$ and $o_2$) of the same 5000 test images of the classes in the first phase of CIFAR-100. The model being evaluated is trained on all the phases of CIFAR-100, i.e., the model obtained after training on phase P=5 of CIFAR-100. Therefore, the classes of the first phase of CIFAR-100 have been previously seen by the model and suffer from catastrophic forgetting. The results indicate that the number of correct predictions is different for the two orientations of the same test images of the previously seen classes. We also report the number of test images that are correctly classified by the model when they have the orientation $o_1$ but not when they have the orientation $o_2$ (i.e., \cmark $o_1$\xmark $o_2$). Similarly, we report the number of test images that are correctly classified by the model when they have the orientation $o_2$ but not when they have the orientation $o_1$ (i.e., \xmark $o_1$\cmark $o_2$). We observe in Table~\ref{tab:difforget}, that a large number of test images are classified correctly by the model when they have the orientation $o_1$ but are classified incorrectly when they have the orientation $o_2$ and vice-versa. We also observe this same behavior for other datasets. This demonstrates that the impact of catastrophic forgetting on the model predictions differs with the orientations of the image. This is a very interesting finding that, to the best of our knowledge, has been obtained for the first time. This motivates us to propose a data-ensemble (EN) approach that combines the predictions for different orientations of the same test image to help the model retain further information regarding the previously seen classes. The results in Table~\ref{tab:difforget} indicate that our data-ensemble (EN) approach achieves a significantly higher number of correct predictions using both orientations of the same test image of phase 1 as compared to using only $o_1$ or $o_2$ orientation of the image. We also observe that our proposed data-ensemble approach correctly classifies all those images (\cmark EN\cmark $o_1$\cmark $o_2$) that are correctly classified by the model for both orientations, i.e. (\cmark $o_1$\cmark $o_2$). Our proposed data-ensemble approach for class-incremental learning can be introduced into any existing class-incremental learning method to improve its performance.

\subsection{Motivation for Dual-Incremental Learning Framework}\label{sec:motivation_dilf}
Directly applying our proposed data-ensemble approach to any incremental learning model severely degrades its performance, as shown in Sec.~\ref{sec:ablnodilf}. The reason behind this problem is that the standard incremental learning methods train the model to identify only the image class labels of the given image and not the orientation classes. As a result, the model performance is very poor for test images with orientations not seen during training. This severely impacts the performance of our data-ensemble approach during the model evaluation. Therefore, we propose a dual-incremental learning framework (DILF) to ensure that the model learns to identify orientation class labels along with the image class labels. During training using DILF, the orientation classes remain the same throughout all the phases, and only new training images are introduced in each phase. This introduces a commonality among the different phases, and the model learns the orientation classes in a \textit{data-incremental} manner in which the orientation classes remain the same across phases. As a result, DILF helps the model to perform well for images with different orientations, thereby improving the effectiveness of the data-ensemble approach. The commonality introduced by DILF across phases can also help the model to learn better representations and reduce the effect of forgetting on the model predictions, which, in turn, will improve its class-incremental learning performance (see Sec.~\ref{sec:gradcam}).

\subsection{Data Modification For a Batch}
\label{dataset_modification}

\begin{figure*}
    \centering
    \includegraphics[width=0.8\textwidth]{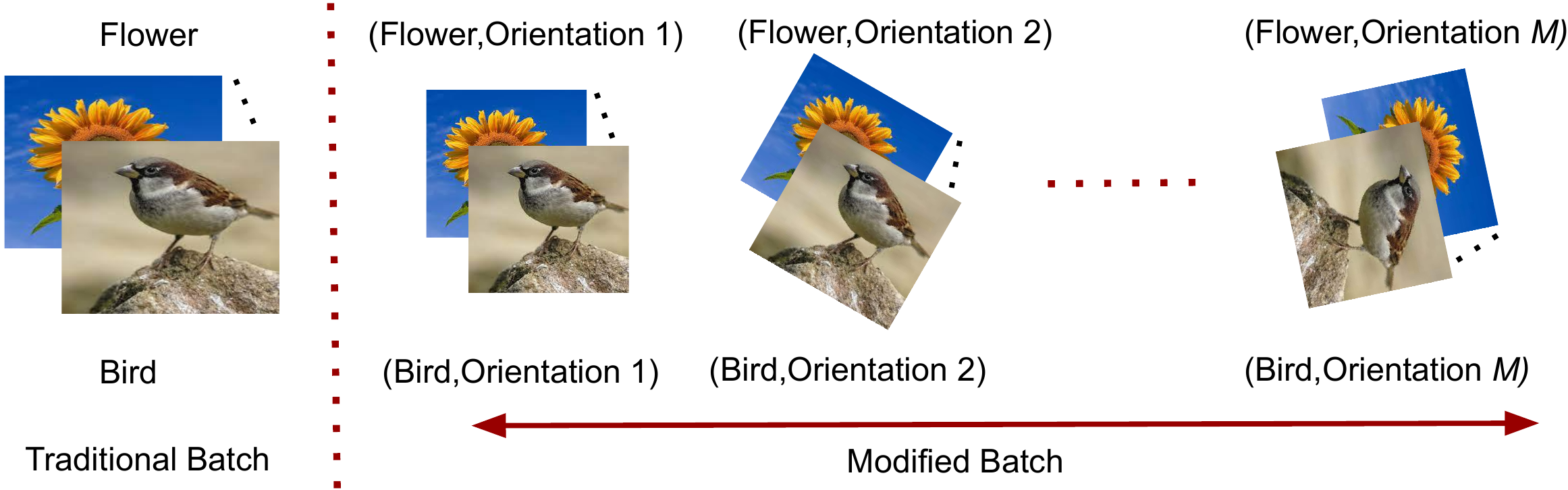}
   
    \caption{The \textbf{data modification step} is performed for each batch in the proposed DILF. Let us assume that each batch originally has $N$ images. We rotate each image (that has a given image class label) with $M$ different orientations in this data modification step. After the data modification, the total number of images per batch increase to $N\times M$, where the label of each image is a tuple of the form (image class, orientation class).}
    \label{batch_modification}

\end{figure*}

Generally, in each phase, training data contains images and the corresponding image class labels. However, DILF has a data modification step wherein, for each image, we also define an orientation class label. A given image is rotated by different orientations belonging to a set of orientation classes, and accordingly, its orientation class label is defined. Let $N$ denote the number of training images in a batch, and $M$ denote the number of orientation classes. The orientation classes remain the same across all phases. Consequently, after the data modification step, the total number of training images in a batch increases to $N\times M$. Furthermore, the label of each image after the data modification is a tuple of the form (image class, orientation class), where the image class is the same as that of the original image (see Figure \ref{batch_modification}). We can also use the training images with different orientations to augment the training data as a data augmentation technique. However, we show in Sec.~\ref{sec:abldass} that our proposed DILF performs significantly better than the data augmentation approach in the class-incremental learning setting.

\subsection{Joint Training of Class and Data incremental Objectives}
\label{joint_training}

\begin{figure*}
    \centering
    \includegraphics[width=\textwidth]{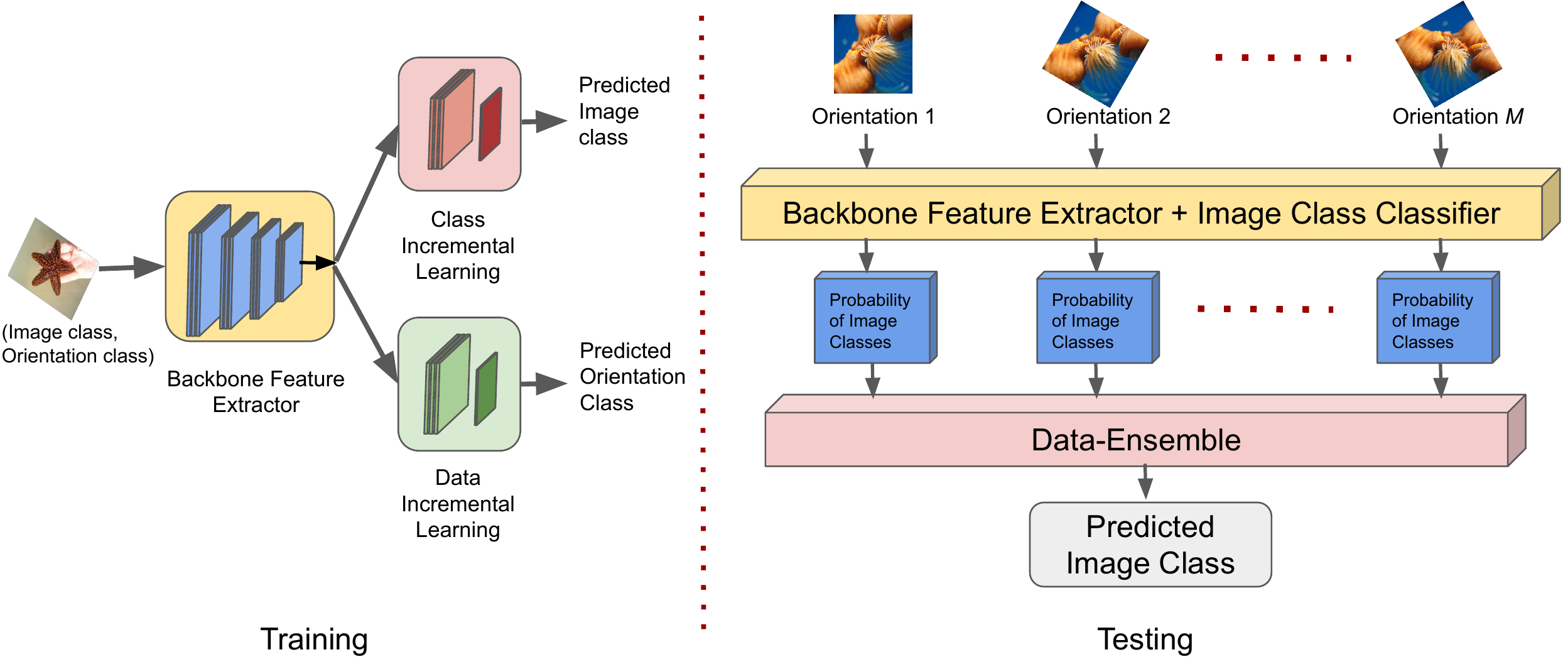}
    \caption{DILF modifies the baseline architecture to introduce commonality across all phases by jointly learning the class and data incremental objectives. During testing, we perform a data-ensemble of differently oriented versions of the same image to predict the image class.}
    \label{framework}
\end{figure*}

As mentioned earlier, we apply DILF to make the model compatible with our proposed data-ensemble approach and improve existing class-incremental learning methods. To implement DILF, the backbone architecture of the baseline class-incremental learning method is modified by adding an additional branch for predicting the orientation class (see training part in Figure \ref{framework}). As a result, the backbone feature extractor ($f_{b}$ with parameters $\bm{\theta_{b}}$) is followed by two classifiers. The first classifier with parameters $\bm{\theta_{i}}$, represented as $f_{i}(\cdot;\bm{\theta_{i}})$ predicts the image class. The second classifier with parameters $\bm{\theta_{o}}$, represented as $f_{o}(\cdot;\bm{\theta_{o}})$ predicts the orientation class (details on the architectures of $f_{i}$ and $f_{o}$ are provided in the Sec.~\ref{sec:netarch}). The set of orientation classes is defined as $\{o_{j}\}_{j=1}^{M}$. Let us assume that at phase $p$, a given training sample $\bm{x}$, rotated by orientation $o_{j}$, is represented as $\bm{\hat{x_j}}$. The feature extracted for image $\bm{\hat{x_j}}$ by the backbone network is denoted by $\bm{\hat{z_j}}$, where $\bm{\hat{z_j}}= f_{b}^{p}(\bm{\hat{x_j}};\bm{\theta_{b}})$. The image class and orientation class predicted at phase $p$ are $f_{i}^{p}(\bm{\hat{z_j}};\bm{\theta_{i}})$ and $f_{o}^{p}(\bm{\hat{z_j}};\bm{\theta_{o}})$ respectively. The corresponding ground truth image class and orientation class are denoted by $y$ and $j$, respectively. Let $\bm{x^{rep}}$ refer to an image in the set of exemplars for the previously seen classes used for replay. The corresponding ground truth image class is denoted by $y^{rep}$. The feature extracted for image $\bm{x^{rep}}$ by the backbone network is denoted by $\bm{z^{rep}}$, where $\bm{z^{rep}}= f_{b}^{p}(\bm{x^{rep}};\bm{\theta_{b}})$. We train the backbone feature extractor $f_{b}$ and the classifier $f_{i}$ on the image class classification task over the modified training images of the current phase using the $\mathcal{L}_{img}$ loss defined below. We train the backbone feature extractor $f_{b}$ and the classifier $f_{o}$ on the orientation class classification task over the modified training images of the current phase using the $\mathcal{L}_{or}$ loss defined below. We also train the backbone feature extractor $f_{b}$ and the classifier $f_{i}$ to prevent catastrophic forgetting using the loss term defined for this objective by the baseline class-incremental learning method.
\begin{equation}
    \mathcal{L}_{img}(\bm{x},y;\bm{\theta_{b}},\bm{\theta_{i}},p)=\frac{1}{M}\sum_{j=1}^{M} \mathcal{L}_{CE}(f_{i}^{p}(\bm{\hat{z_j}};\bm{\theta_{i}}),y)
    \label{eq_img_loss}
\end{equation}
\begin{equation}\label{eq:forget}
    \mathcal{L}_{frgt}(\bm{x^{rep}},y^{rep};\bm{\theta_{b}},\bm{\theta^{-}_{i}},p)= \mathcal{L}_{FRGT}(f_{i}^{p}(\bm{z^{rep}};\bm{\theta^{-}_{i}}),y^{rep})
\end{equation}
\begin{equation}
     \mathcal{L}_{or}(\bm{x};\bm{\theta_{b}},\bm{\theta_{o}},p)=\frac{1}{M}\sum_{j=1}^{M} \mathcal{L}_{CE}(f_{o}^{p}(\bm{\hat{z_j}};\bm{\theta_{o}}),j)
\end{equation}
where  $\mathcal{L}_{CE}$ denotes the cross-entropy loss and $\mathcal{L}_{FRGT}$ is the loss term defined by the baseline method to prevent catastrophic forgetting. In Eq.~\ref{eq:forget}, $\bm{\theta^{-}_{i}}$ refers to using the logits of $\bm{\theta_i}$ corresponding to the only the previously seen classes introduced till the previous phase. Please note that at a given phase $p$, $\mathcal{L}_{img}$ and $\mathcal{L}_{frgt}$ are defined for all the image classes seen till phase $p$. The overall loss function is defined as:
 \begin{multline}
     \mathcal{L}_{total}(\bm{x},y;\bm{\theta_{b}},\bm{\theta_{i}},\bm{\theta_{o}},p)=\mathcal{L}_{img}(\bm{x},y;\bm{\theta_{b}},\bm{\theta_{i}},p)\\+\mathcal{L}_{frgt}(\bm{x^{rep}},y^{rep};\bm{\theta_{b}},\bm{\theta^{-}_{i}},p)\\+\gamma\hspace{1mm}  \mathcal{L}_{or}(\bm{x};\bm{\theta_{b}},\bm{\theta_{o}},p)
     \label{eq_dilf_total_loss}
 \end{multline}
 where $\gamma$ is a hyper-parameter that controls the contribution of $\mathcal{L}_{or}$ in $\mathcal{L}_{total}$. Let $\bm{\hat{\theta_{b}}},\bm{\hat{\theta_{i}}},\bm{\hat{\theta_{o}}}$ represent the parameters of $f_b,f_i,f_o$, respectively, learnt after training. As evident from the training process of DILF, our proposed DILF combines multi-task learning, self-supervision, and data augmentation for training the model. We also discuss multiple approaches for training the network on images having an image class label and an orientation class label in Sec.~\ref{sec:abldass} and experimentally demonstrate that our DILF approach outperforms the other approaches.

\subsection{Data-Ensemble for Model Evaluation}
\label{testing}
We propose a \textit{data-ensemble} approach for performing evaluation in our proposed approach at test time. The term data-ensemble is inspired by ensemble methods. Ensemble methods create multiple variants of the baseline model and combine the predictions from these multiple models (for the same input image) in order to make the final prediction. On the other hand, our proposed data-ensemble creates multiple variants of the input image and combines the predictions for these image variants (using the same model) to make the final prediction (see the testing part in Figure \ref{framework}). Formally, in the proposed data-ensemble approach, during testing at phase $p$, the given input image $\bm{x}$ is rotated according to the $M$ orientation classes to create $M$ versions ($\{\bm{\hat{x_j}}\}_{j=1}^{M}$) of that image. For each of the $M$ versions pertaining to a given test input image, the model outputs the probability of each image class, and the class-wise average of the predicted image class probabilities is used to make the final prediction.

\begin{equation}
    P_{de}(\bm{x})=\frac{1}{M}\sum_{j=1}^{M}f_{i}^{p}(f_{b}^{p}(\bm{\hat{x_j};\bm{\hat{\theta_{b}}}});\bm{\hat{\theta_{i}}})
\end{equation}

where $P_{de}$ denotes the average probability of each image class for the given test input obtained after the data-ensemble. The class with the highest average probability in $P_{de}$ is predicted as the class of that test image. We also perform ablation experiments in Sec.~\ref{ensemble} to validate our data-ensemble approach. 

\section{Experiments}

\subsection{Experimental Setup}

We evaluate the proposed \textit{dual-incremental learning} framework (DILF) on the CIFAR-100 \cite{krizhevsky2009learning} and ImageNet-100 datasets. ImageNet-100 dataset is a subset derived from the ImageNet dataset \cite{russakovsky2015imagenet} as suggested in \cite{rebuffi2017icarl}. DILF and the baseline class-incremental learning methods \cite{hou2019learning,douillard2020podnet,liu2021adaptive,liu2020mnemonics} are evaluated as described in \cite{hou2019learning}. 
For both these datasets, the network is pre-trained with 50 randomly chosen classes. Afterwards, new classes are introduced in an incremental manner. The remaining 50 classes are incrementally added in different phases. If the number of phases is $P$, then each phase has $\frac{50}{P}$ non-overlapping classes. After training on each phase, the model is evaluated on \textit{all the classes seen so far}. For a fair comparison with baseline class-incremental learning methods \cite{hou2019learning,douillard2020podnet,liu2021adaptive,liu2020mnemonics}, we use the same backbone architectures as used by them, i.e., ResNet-32 and ResNet-18 architectures for CIFAR-100 and ImageNet-100 datasets, respectively. For all the compared methods, the \textit{average incremental accuracy} \cite{rebuffi2017icarl} is reported by calculating the average of the classification accuracy obtained after each phase of training. Experiments are repeated three times, and the mean of the average incremental accuracy is reported. The choice of optimizer and the hyper-parameters such as learning rate, memory budget (for replay) per class, etc., are the same as used by the baseline methods. We also use the same exemplar selection strategy as the parent approaches for a fair comparison. Further, we only use the original images from each phase to select the exemplars for data-replay and do not store the transformed images with different orientation classes. The image transformations are carried out on the run, during the training process of each phase. We use \{0,90\} as the set of orientation classes ($M=2$) unless stated otherwise. We analyze different orientation classes for our approach in Sec.~\ref{orientation_classes} and validate the choice of using \{0,90\} as the set of orientation classes.

For all the experiments performed in this research, we follow the same training and testing protocols as defined in the baseline methods (Mnemonics~\cite{liu2020mnemonics}, AANets~\cite{liu2021adaptive}, LUCIR~\cite{hou2019learning}, PODNet~\cite{douillard2020podnet}). For the baseline methods Mnemonics~\cite{liu2020mnemonics} and AANets`\cite{liu2021adaptive}, we use iCaRL \cite{rebuffi2017icarl} and LUCIR \cite{hou2019learning} respectively as the underlying backbone methods. Consequently, we use the ResNet-32 architecture for the experiments on the CIFAR-100 dataset \cite{krizhevsky2009learning} and the ResNet-18 architecture for the experiments on the ImageNet-100 dataset \cite{russakovsky2015imagenet}. Hyper-parameters for all the models are the same as proposed in the baseline methods. We now present the details for the hyper-parameters used in the experiments. For the experiments on the CIFAR-100 dataset, all the models are trained for 160 epochs using the Stochastic Gradient Descent (SGD) optimizer with batch size=128. For training PODNet \cite{douillard2020podnet} on the CIFAR-100 dataset, we use cosine scheduler with initial learning rate= 0.1 and decay factor= 0.1. For training the LUCIR~\cite{hou2019learning}, Mnemonics~\cite{liu2020mnemonics}, AANets~\cite{liu2021adaptive} methods on the CIFAR-100 dataset, the initial learning rate= 0.1 is decayed by factor of 0.1 at epochs 80 and 120. For the experiments on the ImageNet-100 dataset, we train all the models for 90 epochs using the SGD optimizer. For training PODNet on the ImageNet-100 dataset, we use batch size=64, initial learning rate=0.05, and cosine scheduler with a decay factor=0.1. For training the LUCIR~\cite{hou2019learning}, Mnemonics~\cite{liu2020mnemonics}, AANets~\cite{liu2021adaptive} methods on the ImageNet-100 dataset, we use batch size=128 and initial learning rate=0.1 that is decayed by a factor of 0.1 at epochs 30 and 60. For further implementation details, please refer to the corresponding baseline methods \cite{liu2020mnemonics,liu2021adaptive, hou2019learning,douillard2020podnet}. For all the experiments, the memory capacity for replay is fixed to 20 samples per class for the old classes. A detailed analysis on the model performance with different memory capacity (for replay) is presented in Section~\ref{sec:memory}. 

\begin{table*}
	\centering
	\caption{Architecture of orientation classification network of the proposed DILF. W, H, C refer to the width, height and number of channels of the feature maps. Conv refers to a convolutional layer. M refers to the number of orientation classes.}
	\scalebox{1}{
	\addtolength{\tabcolsep}{-4pt}
	\begin{tabular}{ccccc}

        \toprule
		\textbf{Input Size} &\textbf{Output Size} &\textbf{Filter Shape}&\textbf{Stride} &\textbf{Layers}\\

        \midrule
		W$\times$H$\times$C &W$\times$H$\times$C &3$\times$3$\times$C$\times$C  & 1 & Conv + Batch Norm + Leaky\_relu(0.1)\\

		W$\times$H$\times$C &W$\times$H$\times$C &3$\times$3$\times$C$\times$C  & 1 & Conv + Batch Norm + Leaky\_relu(0.1)\\

		W$\times$H$\times$C &W$\times$H$\times$C &3$\times$3$\times$C$\times$C  & 1 & Conv + Batch Norm + Leaky\_relu(0.1)\\

		W$\times$H$\times$C &W$\times$H$\times$C &3$\times$3$\times$C$\times$C  & 1 & Conv + Batch Norm + Leaky\_relu(0.1)\\

	W$\times$H$\times$C &1$\times$1$\times$C & Pool W$\times$H& 1& Average Pool\\

	1$\times$1$\times$C&M&C$\times$M &-& Fully Connected\\

    \bottomrule
	\end{tabular}
	}
		
	\label{orientation_network}
\end{table*}

\subsection{Network Architecture}\label{sec:netarch}
Our proposed DILF has a ResNet backbone feature extractor followed by two branches. The first branch is an image classification network that has an Average Pool layer followed by a Fully Connected (FC) layer. The other branch is the orientation classification network, whose architecture is presented in Table~\ref{orientation_network}. The orientation classification network is common for all phases of incremental learning. For the CIFAR-100 dataset, the value of W (width), H (height), and C (number of channels) are 8, 8, and 64, respectively. For the ImageNet-100 dataset, the value of W, H, and C are 7, 7, and 512, respectively. The feature extracted from the ResNet backbone is forwarded to four convolutional blocks (first four rows of Table~\ref{orientation_network}), followed by Average Pooling and a Fully Connected layer. Each convolutional block constitutes a convolutional layer, a batch normalization layer, and a leaky ReLU activation function with a negative slope of 0.1.

\begin{table*}
\centering
\caption{Comparison of average incremental accuracy obtained on CIFAR-100. Performance of state-of-the-art methods with (w/ DILF-EN) and without introducing proposed DILF-EN are reported. $P$ denotes the number of phases. Baseline results are taken from~\cite{douillard2020podnet,liu2021adaptive,liu2020mnemonics}. $^{\ast}$ denotes methods that do not use replay.}
\label{table_incremental_accuracy_cifar}

	\scalebox{1}{
	\addtolength{\tabcolsep}{-10pt}
\begin{tabular}{lcccc}
 \toprule

  & $P=50$ & $P=25$ & $P=10$ & $P=5$ \\
    \cmidrule{2-5}
 \multicolumn{1}{l}{\ \ \ \ New classes per phase} & 1 & 2 & 5 & 10\\
 \midrule
 LwF$^{\ast}$~\cite{li2017learning} (TPAMI'17) & - & $45.51$ & $46.98$& $49.59$\\
 iCaRL~\cite{rebuffi2017icarl} (CVPR'17) & $44.20$  & $50.60$ & $53.78$& $58.08$\\
 BiC~\cite{wu2019large} (CVPR'19) & $47.09$& $48.96$& $53.21$& $56.86$\\
 TPCIL~\cite{tao2020topology} (ECCV'20) & -& - & $63.58$& $65.34$\\
 \midrule

 LUCIR~\cite{hou2019learning} (CVPR'19) &$49.30$& $57.54$ & $60.14$&$63.17$\\ 
LUCIR \textbf{w/ DILF-EN (Ours)} &\ \ \ \ \ \ \ \ \ $58.66_{\textcolor{purple}{(9.36\uparrow)}}$ & \ \ \ \ \ \ \ \ \ $62.25_{\textcolor{purple}{(4.71\uparrow)}}$ &\ \ \ \ \ \ \ \ \ $65.70_{\textcolor{purple}{(5.56\uparrow)}}$ &\ \ \ \ \ \ \ \ \ $67.38_{\textcolor{purple}{(4.21\uparrow)}}$\\
 \midrule
 Mnemonics \cite{liu2020mnemonics} (CVPR'20) & $53.27$ & $54.13$ &$57.37$ & $60.00$\\
 Mnemonics  \textbf{w/ DILF-EN (Ours)} &\ \ \ \ \ \ \ \ \ \ $57.52_{\textcolor{purple}{(4.25\uparrow)}}$ & \ \ \ \ \ \ \ \ \ \ $59.88_{\textcolor{purple}{(5.75\uparrow)}}$ &\ \ \ \ \ \ \ \ \ \   $62.90_{\textcolor{purple}{(5.53\uparrow)}}$ &  \ \ \ \ \ \ \ \ \ \ $64.71_{\textcolor{purple}{(4.71\uparrow)}}$\\
 \midrule
 PODNet-CNN~\cite{douillard2020podnet} (ECCV'20) &$57.98$ & $60.72$ &$63.19$ & $64.83$ \\
 PODNet-CNN \textbf{w/ DILF-EN (Ours)}  &\ \ \ \ \ \ \ \ \ \ $60.55_{\textcolor{purple}{(2.57\uparrow)}}$ &\ \ \ \ \ \ \ \ \ \  $63.22_{\textcolor{purple}{(2.50\uparrow)}}$ & \ \ \ \ \ \ \ \ \ \  $65.69_{\textcolor{purple}{(2.50\uparrow)}}$ &  \ \ \ \ \ \ \ \ \ \ $67.15_{\textcolor{purple}{(2.32\uparrow)}}$\\
 \midrule
 AANets~\cite{liu2021adaptive} (CVPR'21) &$60.59$ & $63.35$ & $65.66$& $66.74$\\
AANets \textbf{w/ DILF-EN (Ours)}  &\ \ \ \ \ \ \ \ \ \ $\textbf{63.89}_{\textcolor{purple}{(3.30\uparrow)}}$& \ \ \ \ \ \ \ \ \ \ $\textbf{67.63}_{\textcolor{purple}{(4.28\uparrow)}}$ & \ \ \ \ \ \ \ \ \ \ $\textbf{69.21}_{\textcolor{purple}{(3.55\uparrow)}}$ & \ \ \ \ \ \ \ \ \ \  $\textbf{70.77}_{\textcolor{purple}{(4.03\uparrow)}}$  \\
 \bottomrule
\end{tabular}}

\end{table*}

\begin{table*}
\centering
\caption{Comparison of average incremental accuracy obtained on ImageNet-100. Performance of state-of-the-art methods with (w/ DILF-EN) and without introducing proposed DILF-EN are reported. $P$ denotes the number of phases. Baseline results are taken from~\cite{douillard2020podnet,liu2021adaptive}. $^{\ast}$ denotes methods that do not use replay.}
\label{table_incremental_accuracy_imagenet}
\addtolength{\tabcolsep}{0pt}
\scalebox{1}{
\begin{tabular}{@{}lccc}
 \toprule

  & $P=25$ & $P=10$ & $P=5$\\
  \cmidrule{2-4}
 \multicolumn{1}{l}{New classes per phase} & 2 & 5 & 10\\
 \midrule
 LwF$^{\ast}$~\cite{li2017learning} (TPAMI'17) & $44.32$ & $47.64$& $53.62$\\
 iCaRL~\cite{rebuffi2017icarl} (CVPR'17)& $54.56$ & $60.90$  & $65.56$\\
 BiC \cite{wu2019large} (CVPR'19)& $59.65$ & $65.14$  & $68.97$\\
 TPCIL~\cite{tao2020topology} (ECCV'20) & - & $74.81$& $76.27$\\
 \midrule

 LUCIR \cite{hou2019learning} (CVPR'19)  &$61.44$ &$68.32$  &$70.84$\\ 
 LUCIR\textbf{w/ DILF-EN (Ours)} &\ \ \ \ \ \ \ \ \ \ $67.09_{\textcolor{purple}{(5.65\uparrow)}}$ &\ \ \ \ \ \ \ \ \ \ $72.64_{\textcolor{purple}{(4.32\uparrow)}}$  &\ \ \ \ \ \ \ \ \ \ $73.91_{\textcolor{purple}{(3.07\uparrow)}}$ \\
 \midrule
PODNet-CNN \cite{douillard2020podnet} (ECCV'20)&$68.31$ &$74.33$  &$75.54$ \\
 PODNet-CNN \textbf{w/ DILF-EN (Ours)}  &\ \ \ \ \ \ \ \ \ \ $68.94_{\textcolor{purple}{(0.63\uparrow)}}$ &\ \ \ \ \ \ \ \ \ \ $\textbf{76.01}_{\textcolor{purple}{(1.68\uparrow)}}$  &\ \ \ \ \ \ \ \ \ \ $\textbf{78.75}_{\textcolor{purple}{(3.21\uparrow)}}$ \\
 \midrule
 AANets \cite{liu2021adaptive} (CVPR'21)& $67.60$ & $69.22$  &$72.55$ \\
AANets \textbf{w/ DILF-EN (Ours)}  & \ \ \ \ \ \ \ \ \ \ $\textbf{69.02}_{\textcolor{purple}{(1.42\uparrow)}}$ & \ \ \ \ \ \ \ \ \ \  $73.62_{\textcolor{purple}{(4.40\uparrow)}}$  &  \ \ \ \ \ \ \ \ \ \ \  $76.40_{\textcolor{purple}{(3.85\uparrow)}}$\\
 \bottomrule
\end{tabular}}

\end{table*}

\subsection{Classification Performance}
\begin{figure*}
    \centering
    \includegraphics[width=\textwidth]{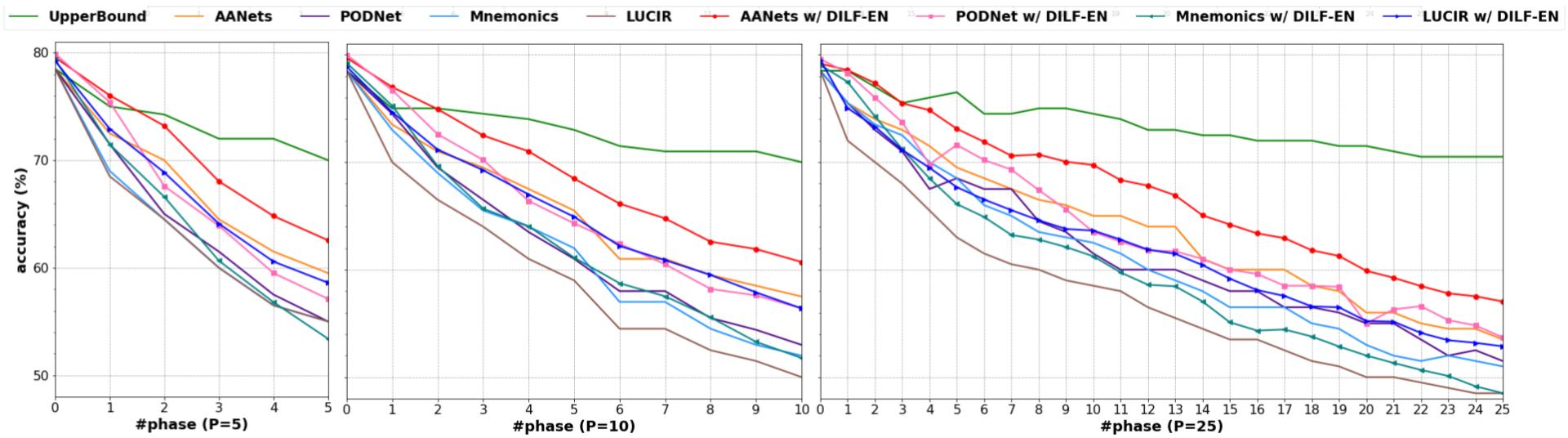}
    \caption{Plot comparing the incremental learning accuracies obtained by incremental learning methods using the proposed DILF-EN (denoted as w/ DILF-EN) on the CIFAR-100 dataset after each phase.}
    \label{fig:cifarplot}
\end{figure*}

Table \ref{table_incremental_accuracy_cifar} and Table \ref{table_incremental_accuracy_imagenet} report the average incremental accuracies obtained for image classification on CIFAR-100 and ImageNet-100  datasets respectively. We observe that the proposed DILF-EN significantly improves the performance of baseline class-incremental learning methods \cite{hou2019learning,douillard2020podnet,liu2021adaptive,liu2020mnemonics} on both the datasets. DILF-EN significantly improves the performance of the best performing state-of-the-art method AANets~\cite{liu2021adaptive} on the CIFAR-100 dataset by an absolute margin of 3.30\%, 4.28\%, 3.55\%, 4.03\%, for the number of phases, $P=50,25,10$ and $5$ respectively. Similarly, on the ImageNet-100 dataset, the performance of AANets is improved by an absolute margin of 1.42\%, 4.40\%, 3.85\%, for $P=25,10$ and $5$ respectively. Figure~\ref{fig:cifarplot} presents the phase-wise accuracies obtained on CIFAR-100 for $P=5,10$ and $25$. We observe that baseline methods trained with DILF-EN achieve the highest accuracies in every evaluated phase. We also provide the accuracy achieved by the LwF~\cite{li2017learning} in this setting for completeness even though it does not use any exemplar memory.

\begin{table}[t]
    \centering
   \caption{Comparison of average incremental accuracy obtained for different approaches for utilizing the modified images with different orientations. The results are reported on CIFAR-100 for $P=5$. DA - Data Augmentation, SS-Self Supervision.}
     \label{tab:dass}
\addtolength{\tabcolsep}{10pt}
\begin{tabular}{lc}
    \toprule
        \textbf{Method}  & \textbf{Accuracy} \\
        \midrule
        AANets (Baseline) & 66.74 \\
        \midrule
        AANets + SS & 66.78 \\
        AANets + DA  & 66.92\\
        AANets \textbf{+ DILF (Ours)}& \textbf{68.14} \\
        \midrule

        AANets + DA + EN & 69.10\\
        AANets + SS + EN & 63.10 \\
        AANets \textbf{+ DILF + EN (Ours)}& \textbf{70.77} \\

        \bottomrule
    \end{tabular}

\end{table}

\subsection{Approaches for Training on Images with different Orientations}\label{sec:abldass}
In our proposed dual-incremental learning framework, we transform each training image to create new images with the same image class but different orientation classes. A naive approach for training the model is to simply use these images as data augmentation (DA), i.e., the model is trained to only predict the image class of these images. The total loss for the data augmentation approach $\mathcal{L}_{total}^{da}$ is defined in Eq.~\ref{eq_aug} below. However, the proposed DILF also trains the model to identify the orientation class of the image using a data-incremental objective (see Eq.~\ref{eq_dilf_total_loss}).

 \begin{equation}
    \mathcal{L}_{total}^{da} = \mathcal{L}_{img}(\bm{x},y;\bm{\theta_{b}},\bm{\theta_{i}},p)+\mathcal{L}_{frgt}(\bm{x^{rep}},y^{rep};\bm{\theta_{b}},\bm{\theta^{-}_{i}},p)
  \label{eq_aug}
\end{equation}

Table~\ref{tab:dass} clearly demonstrates that DILF (AANets + DILF) significantly outperforms the baseline model (AANets). However, the improvement cannot be simply attributed to the increased training dataset as the performance of the baseline model when trained on the augmented dataset (AANets + DA) is marginally higher than the baseline itself. The results also indicate that our DILF-EN approach (AANets + DILF + EN) also significantly outperforms the data augmentation with data-ensemble approach (AANets + DA + EN). Therefore, our approach is significantly better than the data augmentation approach.

Another approach can be to perform orientation prediction as a standard self-supervised (SS) task (see Eq.~\ref{eq_self_supervision}). In this approach, only original images are used for learning the image class classifier, while rotated images are used to train the orientation class classifier as a self-supervised task. The total loss for this approach $\mathcal{L}_{total}^{ss}$ is defined in Eq.~\ref{eq_self_supervision}. Proposed DILF, on the other hand, is trained to jointly learn the orientation and image classes for each image. Consequently, for a given input image, both image class, and orientation class labels are mandatory for training DILF, and the loss is calculated for both the classifiers for each image (see Eqs.~\ref{eq_img_loss}-\ref{eq_dilf_total_loss}).

\begin{multline}
  \mathcal{L}_{total}^{ss}=\mathcal{L}_{CE}(f_{i}^{p}(f_{b}^{p}(\bm{x;\bm{\theta_{b}}});\bm{\theta_{i}}),y)\\+\mathcal{L}_{FRGT}(f_{i}^{p}(f_{b}^{p}(\bm{x^{rep};\bm{\theta_{b}}});\bm{\theta^{-}_{i}}),y^{rep})\\+ \frac{1}{M}\sum_{j=1}^{M} \mathcal{L}_{CE}(f_{o}^{p}(f_{b}^{p}(\bm{\hat{x_j};\bm{\theta_{b}}});\bm{\theta_{o}}),j)
  \label{eq_self_supervision}
\end{multline}

Table~\ref{tab:dass} compares the performance after incorporating orientation prediction as a standard self-supervision task (AANets + SS) and the performance of our proposed DILF (AANets + DILF). We observe that performance obtained by AANets + SS is marginally better compared to the baseline (AANets). On the other hand, AANets + DILF significantly improves AANets and outperforms AANets + SS by a significant margin. The results also indicate that our DILF-EN approach (AANets + DILF + EN) also significantly outperforms the standard self-supervision with data-ensemble approach (AANets + SS + EN). We also observe that the performance of AANets + SS significantly drops on using the data-ensemble (EN), i.e., AANets + SS + EN. This happens because, in the case of SS, the model is not trained to identify the image classes of the transformed images with different orientation classes. Due to this disjoint training, the model is not good at identifying the image class for images with different orientations.

\subsection{Ablation Study}
\subsubsection{Effect of Hyper-parameter $\gamma$}
\label{hyperparameter_gamma}
As reported in Table \ref{table_hyperparameter_gamma}, we find that initially, as the value of $\gamma$ increases, the contribution of data-incremental objective in $\mathcal{L}_{total}$ increases (see Eq.~\ref{eq_dilf_total_loss}) and the average incremental accuracy improves. However, increasing the value of $\gamma$ beyond a threshold reduces the effective contribution of the class-incremental objectives, which adversely affects the average incremental learning accuracy. $\gamma=0.5$ is empirically found to be the best parameter and is used for all the experiments.

\begin{table*}[t]
    \centering
    
    \caption{Average incremental accuracy obtained for different values of hyper-parameter $\gamma$. The results are reported on the CIFAR-100 dataset for AANets w/ DILF-EN for $P=5$.}
    \label{table_hyperparameter_gamma}
    \scalebox{1}{
    \begin{tabular}{cccccccc}
    \toprule
        \textbf{$\gamma$}  &\textbf{$0$}  &\textbf{$0.01$} &\textbf{$0.1$} &\textbf{$0.3$} &\textbf{$0.5$} &\textbf{$0.7$} &\textbf{$1$}\\
        \midrule
       Accuracy &69.23 &69.27 &70.05 &69.73 &\textbf{70.77} &70.41 &69.64\\
        \bottomrule
    \end{tabular}}
    
\end{table*}

\begin{table}[t]
                \caption{Performance for different choices of orientation classes.}
     \label{orientations}
     \addtolength{\tabcolsep}{-0pt}
     \begin{center}
     \scalebox{0.8}{
\begin{tabular}{clcc}
    \toprule
        $M$  &Orientations ($^{\circ}$)&AANets + DILF-EN  & LUCIR + DILF-EN  \\
        \midrule
        1 &\{0\} (Baseline)& 66.74 & 63.17 \\ 
        \midrule
       & \{0,30\} & 69.52 & 66.73\\
       & \{0,45\} & 70.20 & 67.50\\
       & \{0,60\} & 70.72 & 67.47\\
       &\{0,90\} &70.77& 67.38 \\
       &\{0,180\} &68.99& 66.18 \\
       2&\{0,270\} &70.25& 67.53 \\
       \midrule
       &\{0,90,180\} &70.90  &67.57 \\
       &\{0,90,270\}&70.80  &67.75 \\
       3&\{0,180,270\}&70.88  & 67.90\\
        \midrule
     4&\{0,90,180,270\}&69.62  & 66.80\\
    \bottomrule
    \end{tabular}}  
    
     \end{center}
\end{table}

\begin{table}[t]
   \caption{Average incremental accuracy obtained using different data-ensemble strategies: Avg, Mode, and Max. The results are reported on CIFAR-100 for ($P=5$). The baseline method used is AANets~\cite{liu2021adaptive}. EN represents data-ensemble. EN(Avg), EN(Mode), and EN(Max) represent the data-ensemble approaches that use the Avg, Mode, and Max strategies, respectively.}
     \label{data_ensemble}
      \begin{center}
      \scalebox{1}{
\begin{tabular}{ccc}
    \toprule
        Method & M=2 & M=4 \\  
        \midrule
        AANets + DILF &68.14 &66.06 \\
        AANets + DILF + EN(Mode)  &68.87  &68.99 \\
        AANets + DILF + EN(Max) &70.51 &69.12 \\
        AANets + DILF + EN(Avg) \textbf{(Ours)} &\textbf{70.77}&\textbf{69.62} \\
        \bottomrule
    \end{tabular}}
    \end{center}
\end{table}

\subsubsection{Choice of Orientation Classes}
\label{orientation_classes}
We experiment with different orientation classes over two different baseline models (see Table \ref{orientations}). The first observation is that irrespective of the choice of orientation classes, a significant boost in accuracy is observed as compared to the baseline ($M=1$). We observe that for $M=2$, the performance for \{0,90\} is either the maximum or very close to the maximum. Additionally, 0, 90, 180, or 270 are more preferable choices of orientation as compared to 30, 45, or 60 since the former choices can be easily implemented using transpose and flip operations, which results in lower complexity and computational cost. Further, the performance of the model for the 30, 45, 60 orientations are comparable to that of 0, 90, 180, 270. Therefore, we ignore the 30, 45, 60 orientation classes for the rest of the experiments. We also observe that the accuracy obtained for $M=3$ is marginally better than that obtained for $M=2$. On increasing $M$ to 4, the accuracy drops, which shows that simply increasing the number of orientation classes ($M$) does not directly translate to improved performance. This may be caused due to the increased complexity introduced by a higher value of $M$. Please note that using a high value of $M$ increases the computational overheads. Therefore, considering the above analysis, we choose \{0,90\} as the set of orientation classes for all the experiments using our approach.

\subsubsection{Comparison of Different Ensemble Methods}
\label{ensemble}
In Sec.~\ref{testing}, we propose to perform data-ensemble while testing a model after it has been trained using DILF. We experiment with three strategies for data-ensemble, i.e., Avg, Mode, and Max. The Avg strategy obtains the class-wise mean of the image class probabilities for the different orientations of the same image and predicts the image class having the highest mean probability as the output class. In the Mode strategy, the image class label, which has the maximum probability for most of the orientations of the same image, is predicted as the output class. In the Max strategy, the image class label, which has the maximum probability across the different orientations of the same image, is predicted as the output class. Table \ref{data_ensemble} reports that irrespective of the choice of any of the three data-ensemble strategies, the average incremental accuracy obtained without data-ensemble (AANets + DILF) is lesser compared to the accuracy obtained after data-ensemble.  We experiment with two different choices of orientation classes $M=2$ (orientation class set $\{0,90\}$) and $M=4$ (orientation class set $\{0,90,180,270\}$). Avg turns out to be a better data-ensemble technique than Mode and Max for both the choices of orientation classes. Therefore, we use Avg as a data-ensemble technique in all the experiments.

\begin{table}[t]
    \centering
                \caption{Ablation experiments to verify whether DILF is a pre-requisite step for applying data-ensemble (EN). We report the average incremental accuracy obtained by each approach on CIFAR-100 for $P=5$.}
     \label{tab:nodilf}

     \addtolength{\tabcolsep}{10pt}
\begin{tabular}{lc}
    \toprule
        \textbf{Method}  & \textbf{Accuracy} \\
        \midrule
        AANets & 66.74\\
        AANets + EN & 60.74\\
         AANets + \textbf{DILF-EN (Ours)}  & \textbf{70.77} \\

        \bottomrule
    \end{tabular}

\end{table}

\subsubsection{DILF as a pre-requisite for Data-Ensemble}\label{sec:ablnodilf}

In this section, we perform ablation studies to verify whether our proposed dual-incremental learning (DILF) is a pre-requisite step for using data-ensemble. We observe in Table~\ref{tab:nodilf} that applying data-ensemble to AANets without training it using DILF significantly degrades the average incremental accuracy of the model by an absolute margin of 6\% w.r.t. the original AANets model. This occurs because the model has not been trained on images with different orientations, and therefore model performs poorly for images with different orientations. Consequently, the model predictions for images with orientations not seen during training are very poor, and combining the bad predictions for the modified images makes the data-ensemble predictions very poor. Therefore, we cannot apply data-ensemble to a model without first training it using DILF. Otherwise, it will significantly degrade the model performance. 

\subsubsection{Model Performance with Different Memory Budget}
\label{sec:memory}
We also evaluate the performance of DILF-EN under different memory budgets. For these experiments, we compute the performance of AANets with and without incorporating the proposed DILF-EN. Table~\ref{table_memory} reports the average incremental accuracy for these experiments. Intuitively, increasing the number of replay images per class should reduce overfitting and improve the average class incremental accuracy. As expected, we observe that the average incremental accuracy increases as the number of replay images per class increases. Furthermore, we observe that irrespective of the number of image samples per class reserved for replay, DILF-EN improves the classification performance of the baseline model.

\begin{table}[t]
    \centering
    \caption{Average incremental accuracy obtained for different number of replay images per class. The results are reported on the CIFAR-100 dataset for AANets w/ DILF-EN for $P=5$.}
    \label{table_memory}
    \scalebox{0.9}{
    \addtolength{\tabcolsep}{-3pt}
    \begin{tabular}{cccccc}
    \toprule
        \textbf{Replay Images per Class} &\textbf{$5$}  &\textbf{$10$} &\textbf{$50$} &\textbf{$100$} &\textbf{$200$}\\
        \midrule
       AANets &64.83 &65.03 &68.25 &68.76 &69.84\\
       AANets w/ DILF-EN &\textbf{66.10} &\textbf{68.10} &\textbf{71.16} &\textbf{72.34}&\textbf{73.00}\\
    \bottomrule
    \end{tabular}}
\end{table}
\begin{figure*}[!t]
\centering
\includegraphics[width=0.32\textwidth]{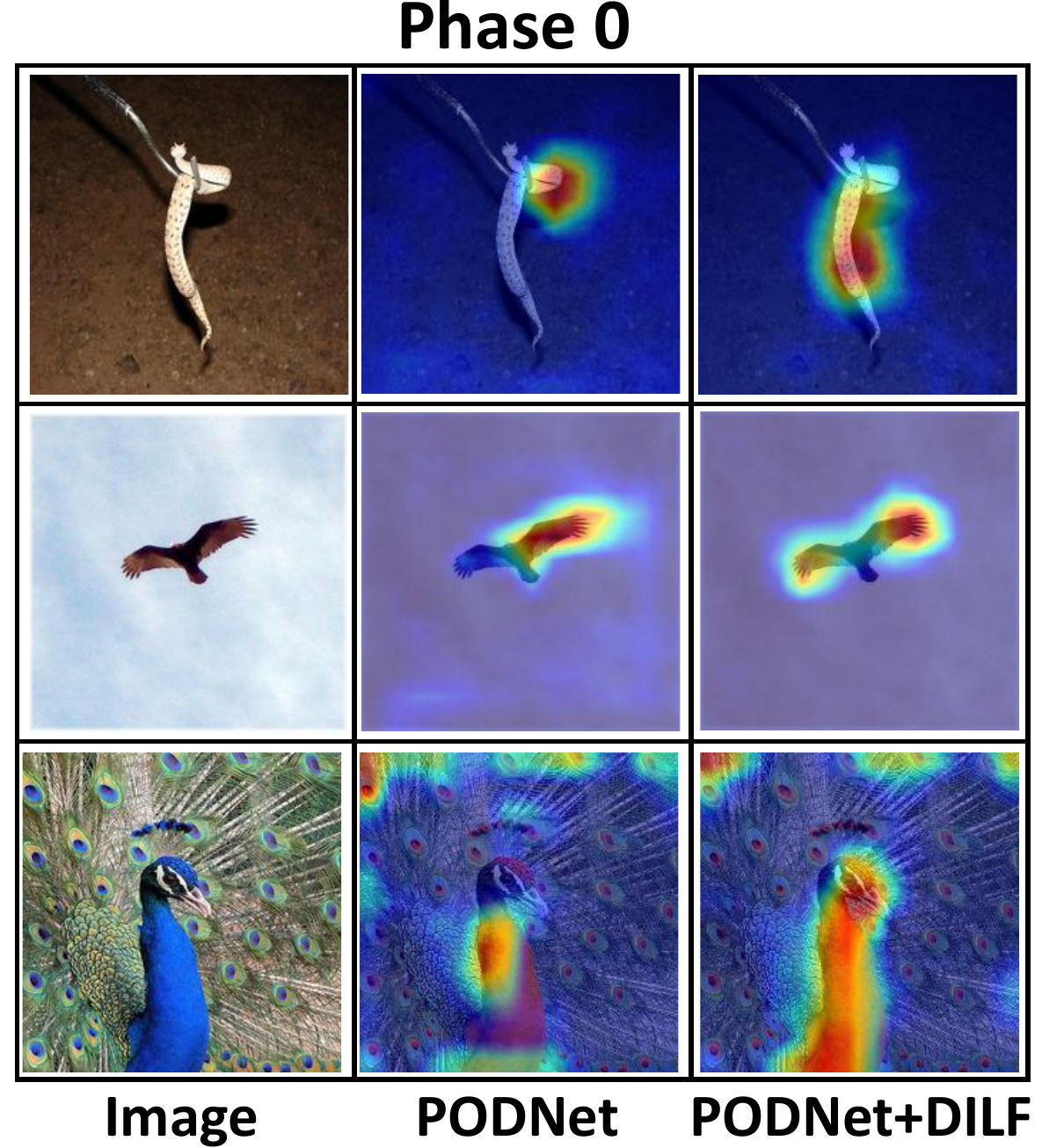}
\includegraphics[width=0.318\textwidth]{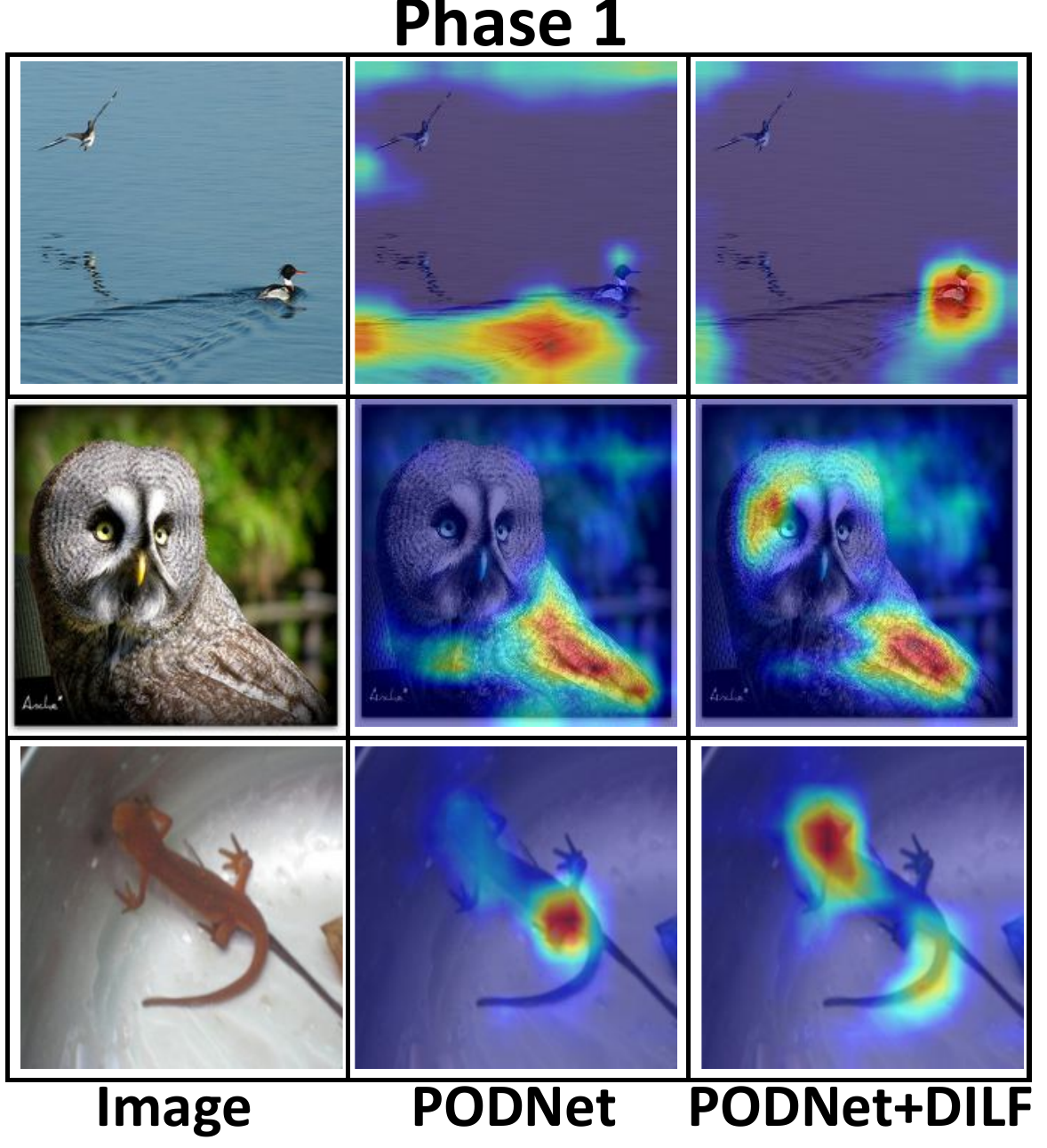}
\includegraphics[width=0.32\textwidth]{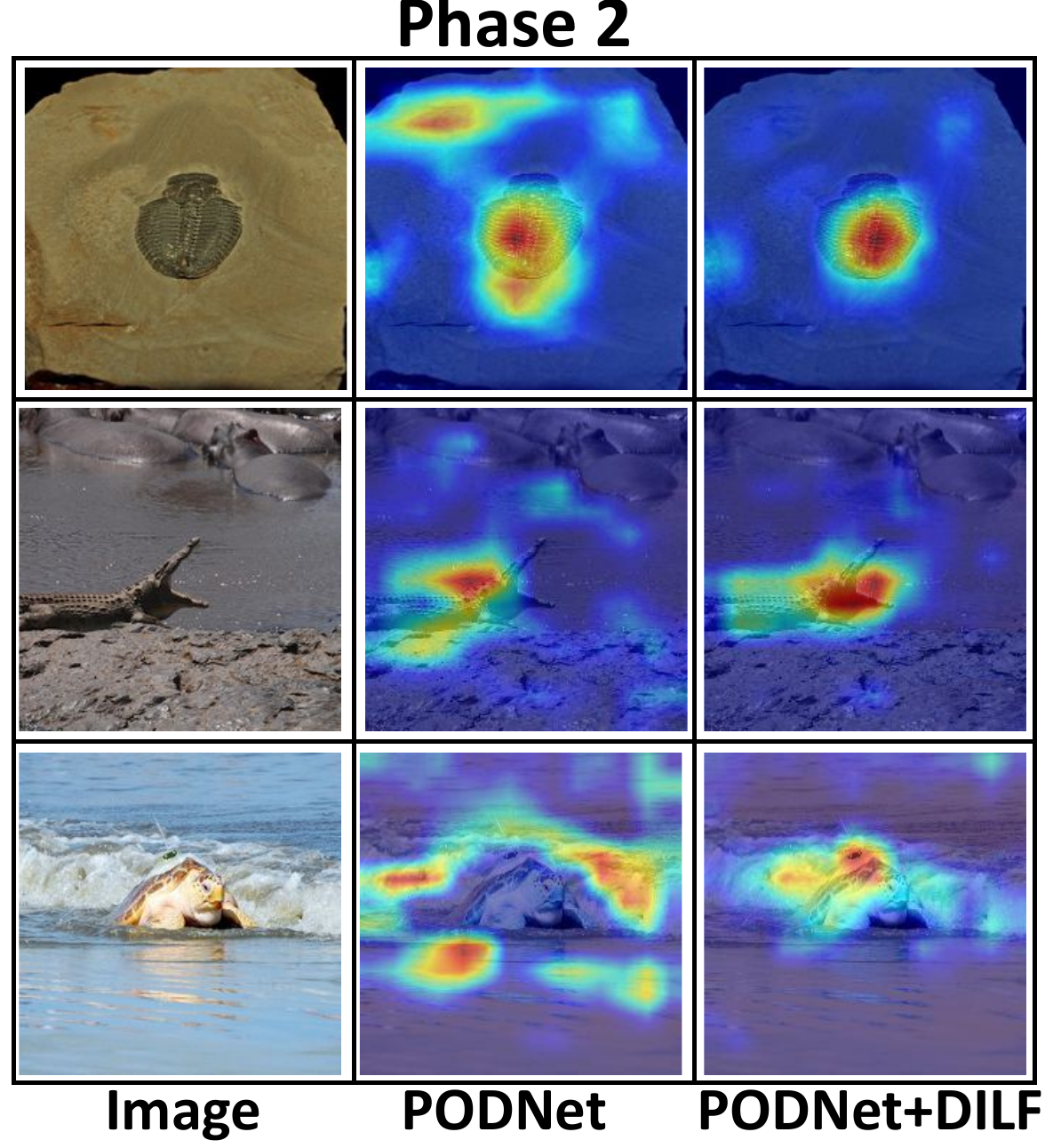}
    \caption{Visualization of the activation maps for the test images from the phases 0, 1 and 2 obtained by applying Grad-CAM \cite{selvaraju2017grad} to the PodNet model trained on the ImageNet-100 dataset (P=5). Please note that we use the final model obtained after completing all phases of training, to visualize the activations for test images from previous phases. Incorporating DILF in PodNet leads to a higher activation of discriminative features (presented in the rightmost column) as compared to the baseline PodNet (presented in the middle column).}
\label{gradcam_visualizations}
\end{figure*}

\subsubsection{Grad-CAM Visualizations}\label{sec:gradcam}
Our proposed DILF introduces commonality among the phases of incremental learning. This commonality also improves the model's capacity to retain the knowledge about the image classes from the older phases. To reaffirm this, we compute the visualization of activation maps for PODNet \cite{douillard2020podnet} trained with and without DILF. The visualizations are obtained using Grad-CAM \cite{selvaraju2017grad}. Figure~\ref{gradcam_visualizations} illustrates the activation maps obtained for the classes seen in the previous phases of training using the PodNet model obtained after completing all phases of training on the ImageNet-100 dataset (P=5), i.e., model obtained after completing the phase 5 training. We observe that the activations of discriminative features are higher after incorporating the proposed DILF as compared to the baseline model. We also observe that the baseline model sometimes obtains high activations on the image regions that are not relevant for the image class classification (see first row of Phase 1, first and third row of Phase 2). Through these sample cases, we find that incorporating DILF helps the model to achieve high activations on relevant image pixels. We also observe that DILF promotes learning of salient features that are sometimes ignored by the baseline model (see third row of Phase 0, second and third row of Phase 1, and second row of Phase 2). These results provide insights on the improved average class incremental accuracy obtained after incorporating the proposed DILF in PodNet.

\section{Conclusion}

In this paper, we propose a novel data-ensemble approach for class-incremental learning that significantly improves the performance of state-of-the-art class-incremental learning methods. The data-ensemble approach applies transformations that change the orientation of a given test image and combines the model predictions for each orientation of that image to improve the class-incremental performance of the model. However, incremental learning models are generally not compatible with the data-ensemble approach. Therefore, we also propose a novel dual-incremental learning framework that jointly trains the model using the class-incremental objective (to learn the image classes) and our proposed data-incremental objective (to learn the orientation classes). We experimentally demonstrate that the model needs to be trained using the dual-incremental learning framework in order to become compatible with the data-ensemble approach. We perform extensive experiments to demonstrate the effectiveness of our approach. We perform various experiments to compare our proposed dual-incremental learning framework with other approaches for training on images with different orientations and empirically demonstrate the superiority of the dual-incremental learning framework. We also perform extensive ablation experiments to validate our approach.

{\small
\bibliographystyle{ieee_fullname}
\bibliography{egbib}
}

\end{document}